\documentclass[11pt,a4paper]{article}
\usepackage{authblk}
\usepackage[hyperref]{acl2019}
\usepackage{times}
\usepackage{latexsym}
\usepackage{url}
\usepackage{amsmath}
\usepackage{natbib}
\usepackage[nameinlink,capitalize]{cleveref}
\usepackage{graphicx}
\usepackage{array}
\usepackage[vlined,linesnumbered]{algorithm2e}

\aclfinalcopy 

\makeatletter
\renewcommand\outauthor{
	\begin{tabular}[t]{>{\centering}p{13cm}} 
		\ifaclfinal 
		\bf\@author
		\else 
		\bf Anonymous ACL submission
		\fi
\end{tabular}}
\makeatother

\title{Demonstration of a Neural Machine Translation System with Online Learning for Translators}

\author[1]{Miguel Domingo}
\author[2]{Mercedes Garc\'ia-Mart\'inez}
\author[2]{Amando Estela}
\author[2]{Laurent Bi\'e}
\author[2]{Alexandre Helle}
\author[1]{\'Alvaro Peris}
\author[1]{Francisco Casacuberta}
\author[2]{Manuel Herranz}

\affil[1]{PRHLT Research Center - Universitat Polit{\`e}cnica de Val{\`e}ncia \protect\\ \{midobal, lvapeab, fcn\}@prhlt.upv.es}
\affil[2]{Pangeanic / B.I Europa - PangeaMT Technologies Division \protect\\ \{m.garcia, a.helle, a.estela, l.bie, m.herranz\}@pangeanic.com}

\date{}

\begin{document}
\maketitle
\begin{abstract}

We introduce a demonstration of our system, which implements online learning for neural machine translation in a production environment. These techniques allow the system to continuously learn from the corrections provided by the translators. We implemented an end-to-end platform integrating our machine translation servers to one of the most common user interfaces for professional translators: {SDL Trados Studio}. Our objective was to save post-editing effort as the machine is continuously learning from human choices and adapting the models to a specific domain or user style.   
\end{abstract}

\section{Introduction}
\label{se:intro}
Productivity is crucial in the translation industry. Nowadays, translation companies must be more competitive than ever and meet fast commercial demands. Thus, they need to produce high quality translations in shorter periods of time. Machine translation (MT) can help them to achieve this goal: instead of a linguist thinking out or ``creating'' translations from scratch, ``humanizing'' automatic translations has become a common process in the industry. This is known as post-editing (PE) and it has been shown to be effective in many cases \citep{Arenas08,Hu16}. As MT systems are continuously improving their capabilities \citep[e.g.][]{Hassan18,Wu16}, this workflow has become of major relevance in the translation industry. Nonetheless, MT technology is still far from perfect \citep{Dale16,Koehn17}, and there is still room for improvement.

Inherently to the PE process, new bilingual data is continuously generated (the post-edited samples). This data is typically used for the creation of domain-specific corpora, useful for adapting systems from a broader domain into a specific domain, client or style. The online learning (OL) paradigm aims to perform this adaptation during the PE process: each time the user validates a post-edited translation, the system is updated as this data is taken into account. Therefore, when the next translation is produced, the system will consider the previous post-editions. It is assumed that better translations (or translations more suited to the human post-editor preferences) will be produced.

The OL paradigm has quickly attracted the attention of researchers and industry. The CasMaCat \citep{Alabau13} and MateCat \citep{Federico14} projects{\textemdash}where phrase-based statistical MT systems were adapted incrementally from the user post-edits{\textemdash}achieved many advances in this direction. More recently, OL techniques were also applied to neural machine translation (NMT) systems \citep{Peris17b,Turchi17,Kothur18,Wuebker18,Peris18}.

In this paper, we introduce a demo system of our in-house OL framework, in which we integrated our translation servers with the translators user-friendly interface SDL Trados Studio.

The rest of this document is structured as follows: \cref{se:ol} introduces the online learning paradigm. Next, in \cref{se:architecture}, we describe in detail our in-house architecture in which this paradigm is implemented. Finally, \cref{se:summary} summarizes the demo system.

\section{Adapting a NMT system via online learning}
\label{se:ol}

We are interested in benefiting from the post-edits generated by the user during the PE process. To that end, we update the system on-the-fly, i.e, as soon as a sentence has been validated by the post-editor. Right after the user confirms a post-edit, we update the models of our NMT system, using the source sentence and the post-edit as a training pair. This adaptation can be done following gradient descent, the regular training method for neural networks.

\section{Architecture}
\label{se:architecture}
Our in-house architecture of the OL framework is composed of three main modules: the MT engine, the user interface and the translation server which links both. 

\begin{figure}[h!]
    \center
    \includegraphics[width=7cm]{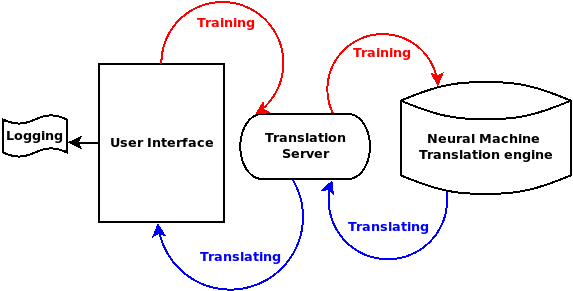}
    \caption{\label{fi:diagram} Architecture of our in-house OL framework. 
        Red arrows represent how the correction made by the user arrives to the Machine Translation engine to be retrained.
        Blue arrows represent how machine translation is delivered to the user.}
\end{figure}

Moreover, we added a logging option to keep user tracking information as keystrokes, time and mouse movements.
\cref{fi:diagram} illustrates this architecture.

The translation process consists of delivering machine translations to the user interface and the training process retrains the MT engine with the feedback provided by the user. Both processes are performed through client-server communication. Next, we describe each module in detail.

\subsection{Machine Translation Engine}

The core of the MT engine is composed by the models generating translations, which can be retrained when required. Each translation project has its own model, whose architecture is set according to the project's need. All models are neural-based and are trained using \texttt{OpenNMT-py} \citep{opennmt}.

Each MT model has its own configuration file, which contains personalized translation and OL options, such as tokenization, subword segmentation, learning rate, etc.

%
%

\subsection{Translation Server}
\label{se:server}
A translation server communicates with the MT models in order to generate translations and adapt the systems based on the user's post editions. This server is based on \texttt{OpenNMT-py}'s REST server and uses the HTTP protocol to define the messages in order to serve user's requests. The code of our translation server is open and available\footnote{\url{https://github.com/midobal/OpenNMT-py/tree/OnlineLearning}}. We created a branch in \texttt{OpenNMT-py} that features this server and is compatible with all its different models.

%
%

The communication between the user interface and the MT engine is performed by means of GET and POST requests. The server waits for translation requests. When received, these requests are sent to the machine translation engine in a JSON format. When a machine translation segment is corrected by the user, the correction is sent to the translation engine.

\begin{figure*}[!ht]
    \center
    \includegraphics[width=13cm]{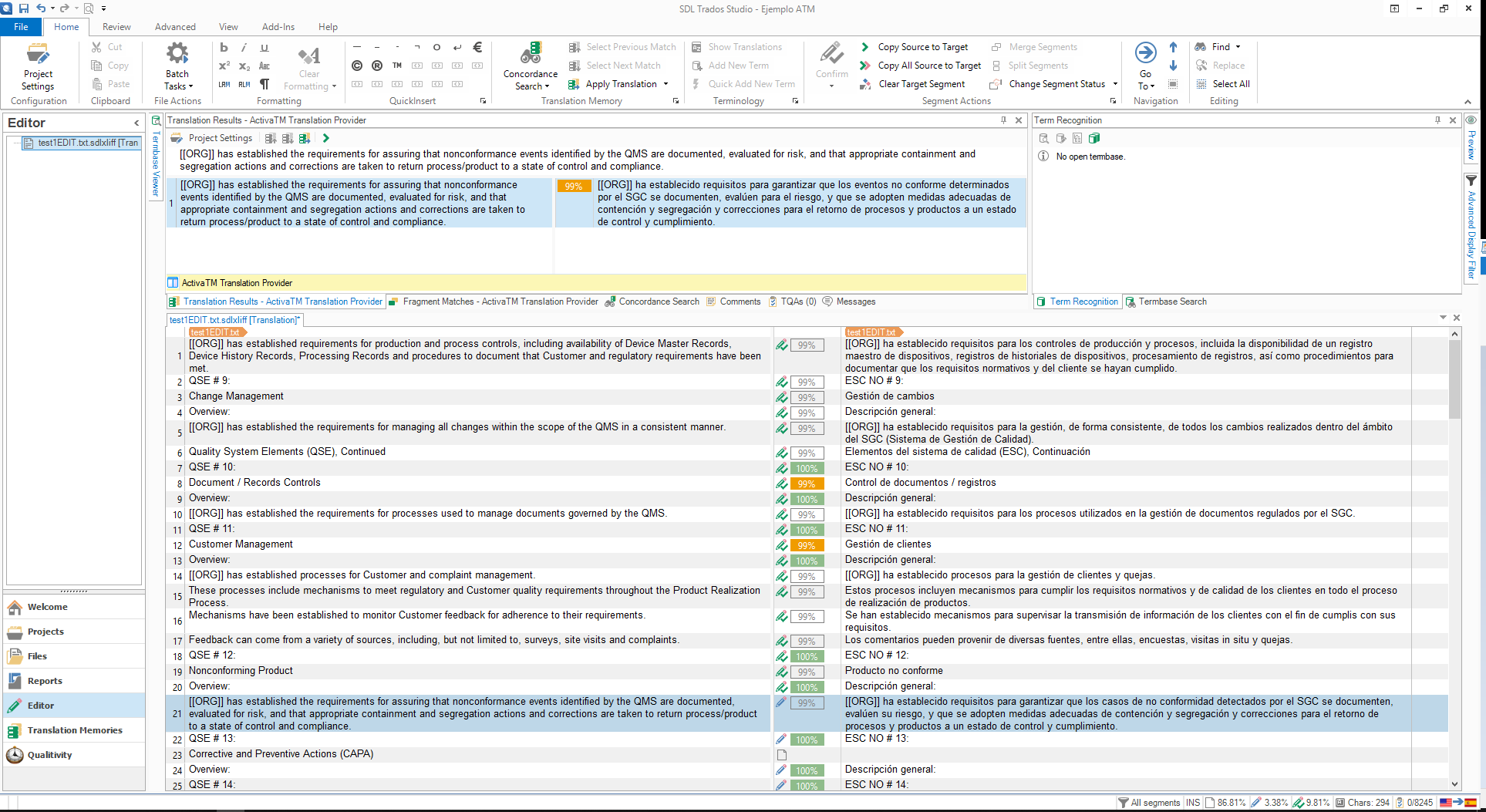}
    \caption{\label{fi:trados} User Interface from Trados Studio SDL.}
\end{figure*}

%


\subsection{User Interface}
\label{se:interface}
In the translation industry, the most common user interface for translators is {SDL Trados Studio}\footnote{\url{https://www.sdl.com/es/software-and-services/translation-software/sdl-trados-studio/}}. \cref{fi:trados} shows the user interface. The user gets the machine translation outputs automatically when the target part of the segment in the interface is clicked.
Then, the user post-edits the segment and, when the translation is corrected, confirms it.

SDL allows the development of plugins for Trados Studio to enhance and extend the tool. Moreover, it has a large developer community\footnote{\url{https://community.sdl.com/developers-more/developers/language-developers}} helping the software with add-ons and apps. We incorporated our adaptive framework as a Trados Studio plugin, that connected the user interface with Trados Studio with our translation server. As the user confirms a post-edit, the reviewed segment is sent back to the MT engine to be retrained with this new information.

\begin{figure}[!ht]
    \includegraphics[width=4cm]{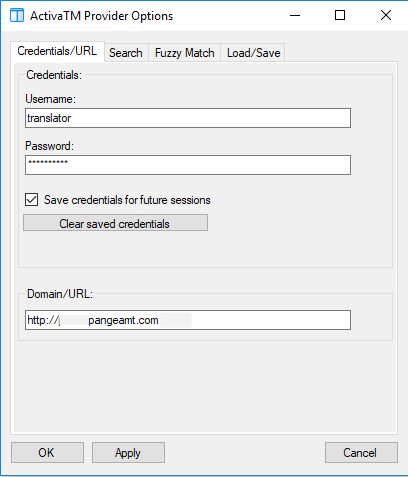}\includegraphics[width=4cm]{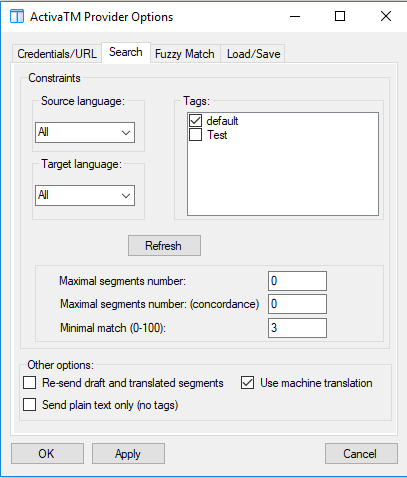}
    \includegraphics[width=8cm]{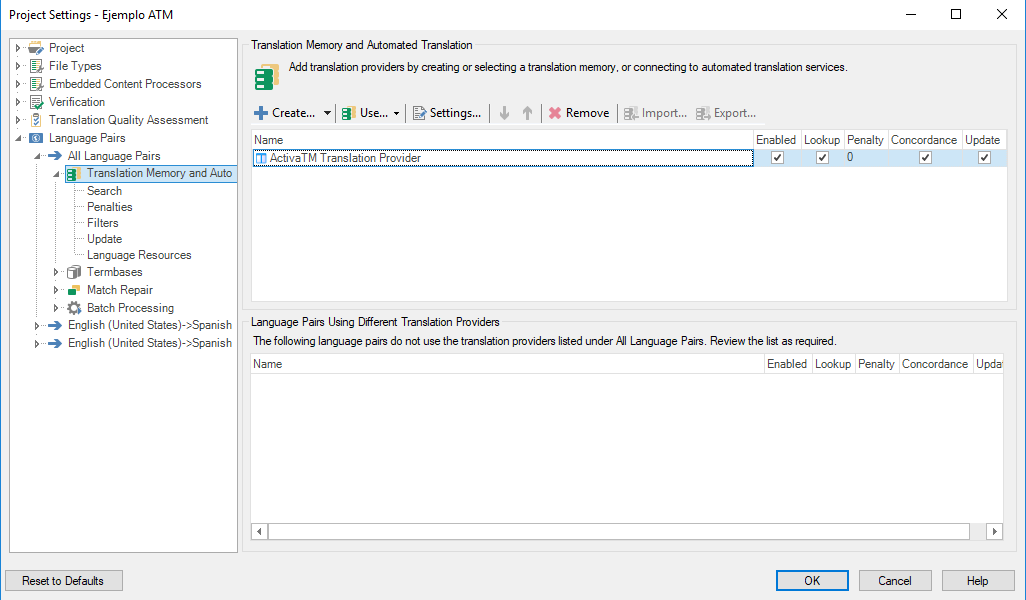}
    \caption{\label{fi:activatm} Machine translation plugin configuration.}
\end{figure}

In order to set up this plugin, the user fills the credentials with a username, password and URL pointing to the server (see \cref{fi:activatm} top left box). 
Also, the user fills the required languages and clicks in the ``use machine translation'' option (see \cref{fi:activatm} top right box).
Finally, all the options have to be enabled in the translation provider plugin (see \cref{fi:activatm} bottom box) in the Trados Studio project settings.

\begin{figure*}[!ht]
    \center
    \includegraphics[width=11cm]{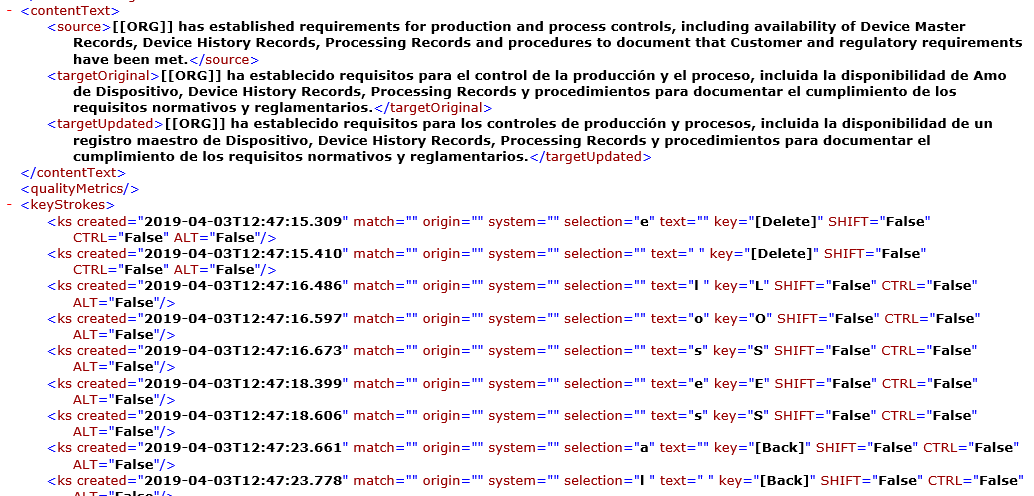}
    \caption{\label{fi:log} Example of {Qualitivity}'s logging file.}
\end{figure*}

\subsection{Logging}
\label{se:logging}

In order to measure the productivity and effectiveness of OL during the PE process, we integrated tools to log the time, keystrokes and mouse movements involved in post-editing a given file. To achieve this, we incorporated the {Qualitivity}\footnote{\url{https://community.sdl.com/product-groups/translationproductivity/w/customer-experience/2251.qualitivity}} plugin for Trados. This plugin generates an XML logging file, which contains all the keystrokes time information per segment.
An example of this logging is shown in Figure \ref{fi:log}.


With all this log information, we can measure the effort required to post-edit a file using MT with OL. Preliminary experiments in simulated and real environments with professional translators \citep{Domingo19}, reported significant improvements of the quality of the translations generated by the MT systems (up to 5.3 points according to hTER, and 7.8 points according to hBLEU), and a significant reduction of the PE time (up to 7.5 second per sentence).

\section{Summary}
\label{se:summary}
We have introduced a demonstration of Pangeanic's translation framework, which incorporates on-the-fly system adaptation via online learning. This paradigm allows human translators /post-editors to produce more human-quality text, that is, be more productive{\textemdash}a fundamental issue in the translation industry{\textemdash}since the system is continuously learning from the user post-edits, avoiding repetition of the same errors.
We have integrated our MT servers into the {SDL Trados Studio} user interface which is one of the most used in the translation industry. This system aims to improve human translators' work by saving time and effort.

\section*{Acknowledgments}
The research leading to these results has received funding from the Spanish Centre for Technological and Industrial Development (Centro para el Desarrollo Tecnol{\'o}gico Industrial) (CDTI) and the European Union through Programa Operativo de Crecimiento Inteligente (Project IDI-20170964). We gratefully acknowledge the support of NVIDIA Corporation with the donation of a GPU used for part of this research.

\bibliography{demo}
\bibliographystyle{acl_natbib}

\end{document}